\documentclass{article}
\usepackage{ijcai25}

\usepackage{times}
\usepackage{soul}
\usepackage{url}
\usepackage[hidelinks]{hyperref}
\usepackage[utf8]{inputenc}
\usepackage[small]{caption}
\usepackage{graphicx}
\usepackage{amsmath}
\usepackage{amssymb}
\usepackage{amsthm}
\usepackage{booktabs}
\usepackage{algorithm}
\usepackage{algorithmic}
\usepackage[switch]{lineno}
\usepackage{enumitem}

\usepackage{algorithm}
\usepackage{algorithmic}
\usepackage{amsmath}
\usepackage{booktabs}
\usepackage{tabularray}
\usepackage{makecell}
\usepackage{array}
\usepackage{multirow}
\usepackage{ragged2e}
\usepackage{booktabs}
\usepackage{amssymb}
\usepackage{newfloat}
\usepackage{listings}
\usepackage{bibentry}
\usepackage{subfigure}
\usepackage{stfloats}
\usepackage{adjustbox}

\renewcommand{\thefootnote}{\fnsymbol{footnote}} 

\title{AdaMixT: Adaptive Weighted Mixture of Multi-Scale Expert \\ Transformers for Time Series Forecasting}

\author{
    Huanyao Zhang${}^{1,2}$   \and
    Jiaye Lin${}^{3}$  \and
    Wentao Zhang${}^{2}$ \and
    Haitao Yuan${}^{3,\dag}$ \And
    Guoliang Li${}^{3,\dag}$ \\
    \affiliations
    ${}^{1}$School of Computer Science, Peking University, Beijing, China \\
    ${}^{2}$Center for Machine Learning Research, Peking University, Beijing, China \\
    ${}^{3}$Tsinghua University, Beijing, China \\
}

\begin{document}
\maketitle

\begin{abstract}
Multivariate time series forecasting involves predicting future values based on historical observations. However, existing approaches primarily rely on predefined single-scale patches or lack effective mechanisms for multi-scale feature fusion. These limitations hinder them from fully capturing the complex patterns inherent in time series, leading to constrained performance and insufficient generalizability. To address these challenges, we propose a novel architecture named Adaptive Weighted Mixture of Multi-Scale Expert Transformers (AdaMixT). Specifically, AdaMixT introduces various patches and leverages both General Pre-trained Models (GPM) and Domain-specific Models (DSM) for multi-scale feature extraction. To accommodate the heterogeneity of temporal features, AdaMixT incorporates a gating network that dynamically allocates weights among different experts, enabling more accurate predictions through adaptive multi-scale fusion. Comprehensive experiments on eight widely used benchmarks, including Weather, Traffic, Electricity, ILI, and four ETT datasets, consistently demonstrate the effectiveness of AdaMixT in real-world scenarios.

\end{abstract}

\section{Introduction}

\footnotetext[2]{Corresponding author.}

Time series forecasting is essential for various fields, aiming to accurately present future values according to historical observations. The rapid advancement of deep learning has spurred significant research in this area, with applications in traffic prediction~\cite{STMGF,POIjn,yuansurvey,MGHSTN}, recommender systems ~\cite{lin2024mitigating,lin2024tlrec}, and weather forecasting~\cite{zhou:2021informer,liu2024timecma,miao2024unified}.

\begin{figure}[t]
\centering
\includegraphics[width=1\linewidth]{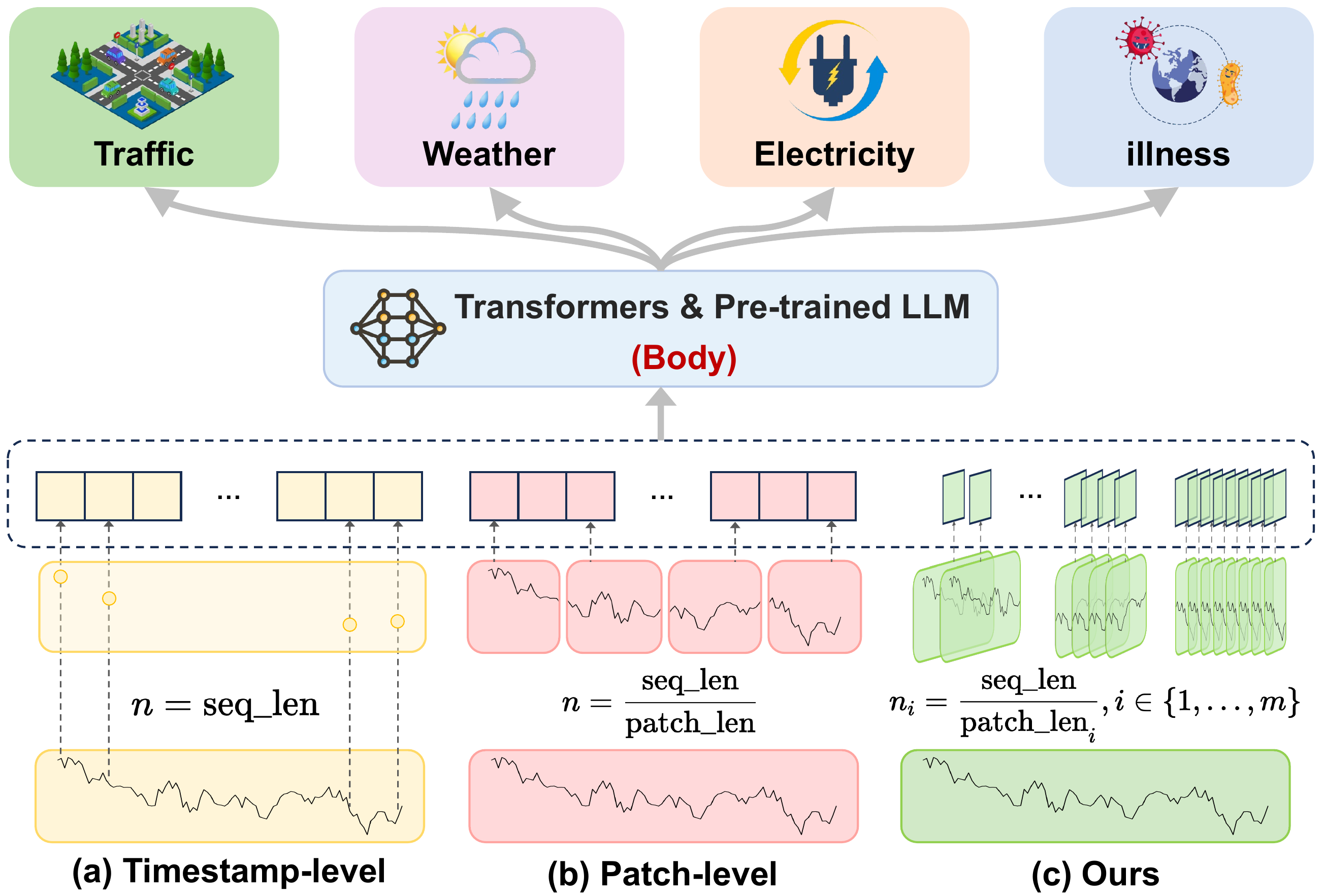}
\caption{\textbf{The existing technologies and our main idea.} (a) Timestamp-level tokenization, where each timestamp is treated as an individual token. (b) Patch-level tokenization, where each time window serves as a token, with $patch\_len$ denoting the length of a patch. (c) Multi-scale feature extraction (Ours), which utilizes $m$ different $patch\_len$ to capture features at varying scales from the time series. In this context, $n$ denotes the number of tokens, and $seq\_len$ refers to the length of time series data.}
\label{fig:1}
\end{figure}

The recent success of attention mechanisms has prompted researchers to investigate the potential of Transformer-based models by redefining time series forecasting tasks as the future token prediction \cite{he2024generalized,RLOMM,DBLP:journals/pvldb/YuanCL24}. Existing research in this field can be categorized into two classes based on different tokenization methods, as illustrated in Figure~\ref{fig:1}. The first approach uses timestamps as tokens, exemplified by models such as Autoformer~\cite{wu:2021autoformer} and FEDformer~\cite{zhou:2022fedformer}. The second approach adopts patch-level tokenization, where a patch represents a window of timestamps, with PatchTST~\cite{nie:2022patchtst} being a notable example. Compared with timestamp-level tokenization, patch-level tokenization more effectively captures temporal patterns, thereby achieving superior forecasting performance. Despite their remarkable success, these methods still face challenges related to limited generalizability. Most existing research focuses primarily on single-scale temporal features, which lack a careful consideration of multi-scale characteristics in time series data. Furthermore, the effective fusion of multi-scale features has not been thoroughly explored, leading to limitations in forecasting under complex scenarios.

To address the above challenges, we propose a novel time series forecasting architecture named Adaptive Weighted Mixture of Multi-Scale Expert Transformers (AdaMixT). By incorporating multi-scale temporal features, AdaMixT significantly enhances both predictive accuracy and generalizability, providing an innovative approach to multi-scale feature fusion. Specifically, AdaMixT leverages both General Pre-trained Models (GPM) and Domain-specific Models (DSM) for time series forecasting, combining the extensive knowledge from GPM with the specialized feature extraction capabilities of DSM to better support downstream tasks. In this framework, input data is segmented into multi-scale patches; smaller patches capture high-frequency features for high-resolution representations, while larger patches capture low-frequency features for low-resolution representations. To effectively integrate multi-scale features, AdaMixT adopts a weighted fusion strategy to assign different weights for the output of each model, generating the final prediction. In summary, we make the main contributions as follows:

\begin{itemize}[leftmargin=0.5cm]
\item We propose a novel multi-scale patch design that combines GPM with DSM, enabling the capture of both short-term and long-term temporal patterns in time series. This integration leverages the open-world knowledge of pre-trained models and the specialized feature extraction capabilities of domain-specific architectures, thereby improving forecasting accuracy and model generalizability.
\item To the best of our knowledge, this work is the first to introduce an adaptive mechanism for multi-scale feature fusion in time series forecasting. By utilizing multiple models as experts and dynamically assigning weights based on the characteristics of time series data, our approach significantly enhances the model’s adaptability to various tasks, ensuring robust scalability across different real-world application scenarios.
\item We conduct extensive experiments across multiple time series forecasting benchmarks, demonstrating that AdaMixT outperforms existing forecasting approaches.
\end{itemize}

\section{Related Work}
\subsection{Time Series Forecasting}
In recent years, considerable research efforts have been directed toward leveraging Transformer-based models for time series forecasting. Early approaches, such as Informer~\cite{zhou:2021informer}, adopt a sequence-to-sequence framework, treating each timestamp as an individual token. Similarly, Autoformer~\cite{wu:2021autoformer} incorporates classical analysis concepts such as decomposition and autocorrelation, while FEDformer~\cite{zhou:2022fedformer} utilizes a Fourier-enhanced structure to achieve linear computational complexity. However, \cite{zeng:2023} highlights the limitations of treating each timestamp as a token, particularly in capturing intricate temporal patterns. To overcome this drawback, models such as Crossformer~\cite{zhang:2023crossformer} and PatchTST~\cite{nie:2022patchtst} are inspired by patch-based visual transformers~\cite{dosovitskiy:2020vit}, representing windows of multiple timesteps as patches and using these as tokens for improved performance. Concurrently, the impressive capabilities of Large Language Models (LLMs) have led to their application in time series forecasting. Models like GPT4TS~\cite{zhou:2023onefit} and TIME-LLM~\cite{jin:2023time-llm} have shown promising results, further demonstrating the potential of LLMs in this field.

Despite these advancements, existing methods~\cite{miao2024less,yuanicde,DBLP:conf/sigmod/Yuan0BF20} still face challenges in effectively capturing and fusing multi-scale features. To overcome these limitations, we propose AdaMixT, which enhances robust multi-scale feature extraction and includes an efficient fusion mechanism. By combining the strengths of GPM and DSM, AdaMixT offers a comprehensive and efficient solution for time series analysis.

\subsection{Multi-scale Feature Learning}
The analysis of multi-scale features plays a pivotal role in numerous fields. In computer vision, multi-scale features enable the extraction of information at varying spatial resolutions within an image, facilitating the analysis of both fine-grained local details and overarching global structures~\cite{das:2020fastcv,he2024mutual}. Meanwhile, multi-scale features have also been widely applied in domains such as knowledge graphs \cite{li2025multi,li2022knowledge}. In recent years, multi-scale analysis has also been increasingly adopted in time series forecasting. For instance, TimesNet~\cite{wu:2022timesnet} transforms one-dimensional sequences into two-dimensional tensors to capture diverse periodic patterns. In addition, MICN~\cite{wang:2023micn} utilizes convolutions with varying kernel sizes to learn features at multiple temporal scales. TimeMixer~\cite{wang2024:timemixer} enhances predictive performance by decomposing multiscale series and effectively blending their seasonal and trend components.

Unlike existing works~\cite{miao2025parameter,DBLP:journals/pvldb/0002WBW23}, AdaMixT learns multi-scale features during the training phase, enabling more efficient feature representation. Furthermore, instead of relying on straightforward feature fusion methods such as concatenation or addition, we design a gating network to score multi-scale temporal features and propose an innovative multi-scale feature fusion mechanism, which further enhances the performance and generalizability.

\subsection{Mixture of Experts}
Mixture of Experts (MoE) refers to a model consisting of different components, known as experts, each specialized in handling distinct tasks or specific aspects of data. Initially introduced in the literature~\cite{jacobs71:1991adaptive}, it has since been extensively studied and refined in subsequent works~\cite{aljundi3:2017expert}. The advent of sparse-gated MoE~\cite{shazeer135:2017outrageously}, particularly within large Transformer-based language models~\cite{lepikhin86:2020gshard}, has revitalized this technique, expanding its applicability and effectiveness. Traditional MoEs typically rely on feedforward networks to select models based on specific scenarios. Unlike these designs, AdaMixT introduces multi-scale feature fusion into time series analysis for the first time. By adopting a self-learning mechanism, AdaMixT focuses on feature fusion rather than model selection, enabling the model to automatically identify critical multi-scale features and achieve efficient integration.

\begin{figure*}[!htbp]
\centering
\includegraphics[width=0.95\linewidth]{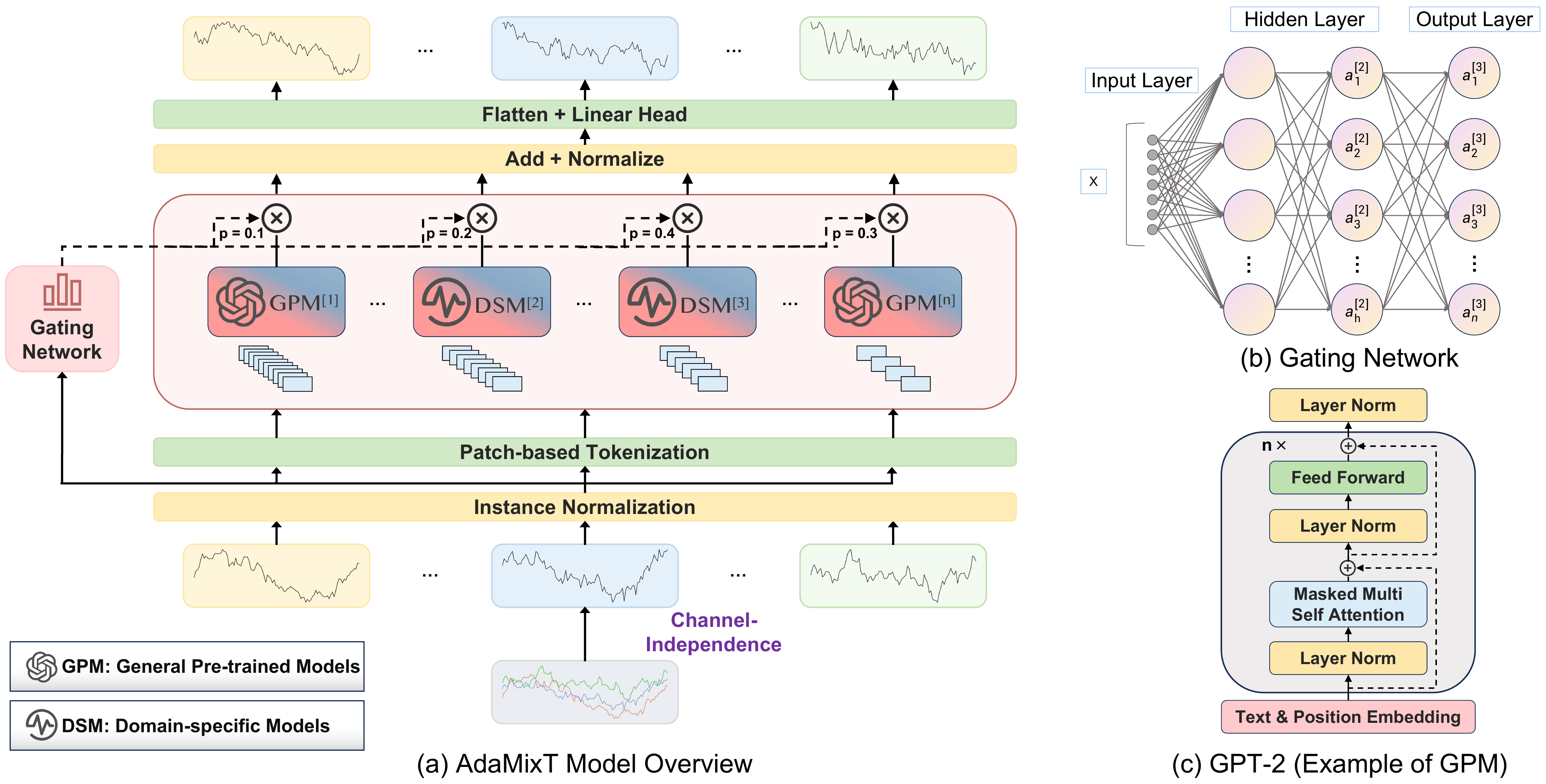}
\caption{\textbf{The architecture of AdaMixT.} (a) AdaMixT adopts a channel-independent design, extracting multi-scale features through patches of varying lengths and feeding these features into multiple experts, including GPMs and DSMs. The outputs of experts are dynamically weighted and fused via a gating network, ultimately generating the final prediction. (b) Gating Network dynamically selects and fuses the outputs of multiple experts, where $X$ denotes the time series features before patching, $h$ denotes the number of hidden neurons, and $n$ denotes the number of experts. (c) GPT-2, serving as a representative of the GPM, can efficiently capture general features and provide a powerful foundational representation for subsequent multi-scale feature fusion.}
\label{fig:2}
\end{figure*}

\section{Method}
In this section, we first provide a formal definition of the problem along with necessary notations and then explain the details of the overall model structure.

\subsection{Problem Definition}
Multivariate time series forecasting involves predicting future values based on historical observation. The objective is to forecast future values for the next $K$ timestamps, given the observations from the previous $L$ timestamps. A multivariate time series comprises multiple related time series, each representing a different variable. Let $\mathbf{x}_{t} = [\mathbf{x}^{(1)}_t, \mathbf{x}^{(2)}_t, \dots, \mathbf{x}^{(M)}_t] \in \mathbb{R}^{M \times 1}$ be a multivariate signal, where $\mathbf{x}^{(i)}_{t}$ denotes the $i$-th variate at time $t$, for $1 \leq i \leq M$. 

In Equation~\eqref{eq:1}, our goal is to develop a model that accurately predicts the values for the next $K$ timestamps based on the recent history spanning $L$ timestamps. Here, $L$ and $K$ are referred to as the look-back window and the prediction horizon, respectively. For a single variable, $\hat{x}_{L+1}^{(i)}, \ldots, \hat{x}_{L+K}^{(i)}$ represents the sequence of future values in the next $K$ time steps for the $i$ th variable, which is the prediction target. In contrast, $x_{1}^{(i)}, \ldots, x_{L}^{(i)}$ denote the observed data for the $i$-th variable over the past $L$ time steps. The function $F$, parameterized by $\Phi$, leverages these observations to effectively forecast the future sequence values.

\begin{equation}
\hat{x}_{L+1}^{(i)}, \ldots, \hat{x}_{L+K}^{(i)} = F\left(x_{1}^{(i)}, \ldots, x_{L}^{(i)}; \Phi\right)
\label{eq:1}
\end{equation}

\subsection{Model Structure}
The architecture of AdaMixT is illustrated in Figure \ref{fig:2}. The model leverages GPMs (e.g., GPT2~\cite{radford:2019gpt2} and Llama~\cite{touvron2023:llama}) and DSMs (e.g., PatchTST~\cite{nie:2022patchtst}) as experts to efficiently extract multi-scale features from time series, thereby achieving accurate prediction objectives. The entire framework consists of three core modules: Multi-Scale Feature Extraction, Expert Pool, and Adaptive Weighted Gating Network (AWGN).

As detailed in Algorithm \ref{alg:AdaMixT}, AdaMixT adopts a channel-independent framework to process each variable. The process begins with instance normalization to standardize each variable. Following this, the time series is segmented into patches of varying lengths using a patch-based tokenization, enabling the capture of multi-scale temporal features spanning both short-term and long-term dependencies. These patches are subsequently processed by multiple experts, including GPMs for general feature representation and DSMs tailored for time-series forecasting. Each expert extracts features from input patches and integrates their outputs based on the scoring results of Adaptive  Weighted Gating Network. The fused results are then passed through Linear Head to generate the final predictions. Furthermore, while AdaMixT inherently employs a channel-independent design, its architecture is highly flexible and can be seamlessly extended to other patch-based time series models, such as Crossformer \cite{zhang:2023crossformer}, for modeling inter-variable dependencies. This versatility establishes AdaMixT as an efficient and generalizable solution for time series forecasting tasks.

\begin{algorithm}[!t]
\caption{Forcasting Process of AdaMixT}
\label{alg:AdaMixT}
\textbf{Input}: Multivariate time series $\mathbf{X}_{1:L}$ with $M$ variables, Look-back window $L$, Prediction horizon $K$ \\
\textbf{Parameters}: Learning rate $\eta$, Batch size $B$, Patch length $P$, Stride $S$, Scale factors $\{F_1, F_2, \dots, F_n\}$\\
\textbf{Output}: Predictions $\hat{\mathbf{X}}_{L+1:L+K}$
\begin{algorithmic}[1]
\FOR{$i \gets 1$ \textbf{to} $M$} 
    \STATE $x\_{norm}^{(i)}_{1:L} \gets$ Instance\_Normalization($x^{(i)}_{1:L}$) 
\ENDFOR

\FOR{$i \gets 1$ \textbf{to} $M$}
    \FOR{$j \gets 1$ \textbf{to} $n$}
        \STATE $p^{(i)}_j \gets$ MF\_Extract($x\_{norm}^{(i)}_{1:L}$, $P \cdot F_{j}$, $S \cdot F_{j}$)
    \ENDFOR
\ENDFOR

\FOR{$i \gets 1$ \textbf{to} $M$}
    \FOR{$j \gets 1$ \textbf{to} $n$}
        \STATE $E^{(i)}_j \gets$ Expert\_Model$_j(p^{(i)}_j)$ \hfill
    \ENDFOR
    \STATE $G^{(i)} \gets$ Gating\_Network($x\_{norm}^{(i)}_{1:L}$)
    \STATE $fused\_feature^{(i)} \gets \sum_{j=1}^{n} G^{(i)}_j \cdot E^{(i)}_j$
\ENDFOR

\FOR{$i \gets 1$ \textbf{to} $M$}
    \STATE $\hat{x}^{(i)}_{L+1:L+K} \gets$ Linear\_Head($fused\_feature^{(i)}$) \hfill 
\ENDFOR
\STATE \textbf{return} $\hat{\mathbf{X}}_{L+1:L+K}$
\end{algorithmic}
\end{algorithm}

\paragraph{Forward Process.} We denote the $i$-th univariate sequence of length $L$ as \( \mathbf{x}^{(i)}_{1:L} = (\mathbf{x}^{(i)}_1 , \dots, \mathbf{x}^{(i)}_L) \), where \( i = 1, \dots, M \). The input sequence \( (\mathbf{x}_1, \dots, \mathbf{x}_L) \) is divided into \(M\) univariate sequences \( \mathbf{x}^{(i)} \in \mathbb{R}^{1 \times L} \), each independently processed by a model consisting of multiple experts, with each sequence using a different patch length. Finally, different weights are assigned to each expert model, and multi-feature fusion is performed. After passing through the final Linear Head, the corresponding prediction results are returned as \( \hat{\mathbf{x}}^{(i)} = (\hat{\mathbf{x}}^{(i)}_{L+1}, \dots, \hat{\mathbf{x}}^{(i)}_{L+K}) \in \mathbb{R}^{1 \times K} \).

\paragraph{Multi-scale Feature Extraction.} Time segments composed of multiple consecutive timestamps are essential for learning effective predictive representations~\cite{nie:2022patchtst}. Building on this idea, we incorporate multi-scale features of time series into the modeling process by dividing each input univariate time series \(\mathbf{x}^{(i)}_{1:L}\) into patches of varying lengths, which may be overlapping or non-overlapping. 
Specifically, we define the base patch length $P$ and stride $S$, and utilize \( Scale\_factors = \{F_1, F_2, \dots, F_n\} \) to adjust the scale sizes, where $n$ denotes the total number of defined scales. Through this patching process, \( n \) patch sequences \(\mathbf{x}^{(i)(j)}_{1:L} \in \mathbb{R}^{P_j \times N_j}\) are generated, where \(
N_j = \left\lfloor \frac{L - P_j}{S_j} \right\rfloor + 2 \) represents the number of patches derived from the \( j \)-th patching operation. Here, \( P_j = P \cdot F_j \) denotes the adjusted patch length, and \( S_j = S \cdot F_j \) represents the adjusted stride at the \( j \)-th scale. To ensure alignment, the original sequence \(\mathbf{x}^{(i)}_{1:L}\) is padded at the end with \( S_j \) repeated values before performing the patching process.

By adopting patch lengths of varying scales, smaller \( P_j \) values enable the  \(\mathbf{x}^{(i)(j)}_{1:L}\)  branch to specifically focus on short-term temporal features with finer granularity, thereby achieving high-resolution modeling. In contrast, larger \( P_j \) values are better suited for effectively capturing long-term seasonal variations and trend characteristics, thus achieving the goal of multi-scale feature extraction.

\paragraph{Expert Pool.}
We propose the concept of an Expert Pool, designed to integrate the strengths of different models to enhance the performance of multivariate time series forecasting. The Expert Pool consists of two core types of models: General Pre-trained Models and Domain-Specific Models. GPMs (e.g., GPT-2~\cite{radford:2019gpt2} and Llama~\cite{touvron2023:llama}) demonstrate exceptional performance in understanding complex time series data due to their strong generalizability and rich feature representation learning. In contrast, DSMs (e.g., PatchTST~\cite{nie:2022patchtst}) focus on precise pattern extraction in time series data, capturing fine-grained features and effectively compensating for the limitations of GPMs in specific tasks.

Unlike existing methods~\cite{jin:2023time-llm,wang:2023micn} that typically utilize either GPM or DSM individually, our approach combines these two types of models, significantly improving prediction performance through the synergistic effects. Both types of models are based on the Transformer architecture, whose core mechanism is the Attention Mechanism. Specifically, patches are first projected into the Transformer latent space of dimension \( D \) using a trainable linear projection \( \mathbf{W}_p \in \mathbb{R}^{D \times P} \), and a learnable positional encoding \( \mathbf{W}_{\text{pos}} \in \mathbb{R}^{D \times N} \) is added to preserve temporal order. The transformed patches are then processed by the multi-head attention module, where \textbf{Query (Q)}, \textbf{Key (K)}, and \textbf{Value (V)} matrices are computed as follows:
\(
\mathbf{Q}_{h}^{(i)} = \left(\mathbf{x}_{d}^{(i)}\right)^{T} \mathbf{W}_{h}^{Q},
\)
\(
\mathbf{K}_{h}^{(i)} = \left(\mathbf{x}_{d}^{(i)}\right)^{T} \mathbf{W}_{h}^{K},
\)
\(
\mathbf{V}_{h}^{(i)} = \left(\mathbf{x}_{d}^{(i)}\right)^{T} \mathbf{W}_{h}^{V},
\)
where $\mathbf{W}_{h}^{Q}, \mathbf{W}_{h}^{K} \in \mathbb{R}^{D \times d_k}$ and $\mathbf{W}_{h}^{V} \in \mathbb{R}^{D \times D}$. The attention output \(
\mathbf{O}_{h}^{(i)} \in \mathbb{R}^{D \times N}:
\) is then computed using a scaled dot-product: 
\(
\left(\mathbf{O}_{h}^{(i)}\right)^{T} = \text{Softmax}\left(\frac{\mathbf{Q}_{h}^{(i)} \mathbf{K}_{h}^{(i)T}}{\sqrt{d_k}}\right) \mathbf{V}_{h}^{(i)}.
\)

\paragraph{Adaptive Weighted Gating Network.} In this study, we proposed AWGN, which differs from traditional simple feature fusion methods such as feature addition or concatenation. This innovative approach effectively considers the importance of different features, thereby enhancing the accuracy of time series forecasting. As shown in Figure~\ref{fig:2}(b), AWGN is a three-layer MLP network that dynamically assigns weights  $G(x^{(i)})$  to the expert models based on the characteristics of each input sequence $x_{(i)}$. These weights are subsequently applied to the output features $E(p^{(i)})$ of the expert models, as described in Equation~\eqref{eq:2}. Here, $x^{(i)}$ represents the time series features before patch partitioning, $p^{(i)}$ denotes the sequence after partitioning, and  $n$  indicates the total number of experts in the expert model pool.
\begin{equation}
y=\sum_{i=1}^{n} G(x^{(i)}) \cdot E(p^{(i)})
\label{eq:2}
\end{equation}
Finally, the features weighted by AWGN are aggregated and passed through Linear Head to generate the final prediction values. This scoring and weighted aggregation mechanism offers a more efficient approach to integrating multi-feature information for time series forecasting tasks.

\paragraph{Loss Function.} We utilize the Mean Squared Error (MSE) loss to quantify the discrepancy between the predictions and the ground truth. The loss in each channel is gathered and averaged over $M$ time series to get the overall objective loss.
\begin{equation}
\mathcal{L}_{\text{MSE}} = \mathbb{E}_{x} \frac{1}{M} \sum_{i=1}^{M} \left\|\hat{x}^{(i)}_{t:t+K-1} - x^{(i)}_{t:t+K-1} \right\|_{2}^{2}
\label{eq:3}
\end{equation}

\paragraph{Instance Normalization.} This technique has been recently introduced to address the distribution shift between training and testing data \cite{ulyanov:2016instance}. It works by normalizing each time series instance $x^{(i)}$ to have a zero mean and unit standard deviation. Specifically, each $x^{(i)}$ is normalized before patching, and the mean and standard deviation are added back to the output prediction afterward.

\section{Experiments}
\subsection{Experimental Setup}
\paragraph{Dataset.} 
We conduct experiments on eight widely recognized benchmarks, namely Weather, Traffic, Electricity, ILI, and four ETT datasets (ETTh1, ETTh2, ETTm1, and ETTm2), which were originally introduced by~\cite{wu:2021autoformer}. The statistics are summarized in Table~\ref{tab:1}.

\begin{table}[!t]
\centering
\renewcommand{\arraystretch}{1.2}
\resizebox{0.48\textwidth}{!}{%
\begin{tabular}{ccccccccc}
\toprule
\textbf{Dataset} & \textbf{\# Features} & \textbf{\# Timesteps} & \textbf{Frequency} & \textbf{Domain}\\
\midrule
Weather & 21 & 52696 & 10 min & Weather\\
Traffic & 862 & 17544 & 1 hour & Transportation\\
Electricity & 321 & 26304 & 1 hour & Electricity\\
ILI & 7 & 966 & 1 week & Illness\\
ETTh1 & 7 & 17420 & 1 hour & Temperature\\
ETTh2 & 7 & 17420 & 1 hour & Temperature\\
ETTm1 & 7 & 69680 & 15 min & Temperature\\
ETTm2 & 7 & 69680 & 15 min & Temperature\\
\bottomrule
\end{tabular}
}
\caption{Statistics of datasets in various domains.}
\label{tab:1}
\end{table}
\paragraph{Baselines.} 
We evaluate AdaMixT against ten SOTA baselines across four categories: (i) LLM-based models, including TIME-LLM~\cite{jin:2023time-llm} and GPT4TS~\cite{zhou:2023onefit}; (ii) Multi-scale time series models, such as TimeMixer~\cite{wang2024:timemixer}, MICN~\cite{wang:2023micn}, and TimesNet~\cite{wu:2022timesnet}; (iii) Transformer-based models, including iTransformer~\cite{liu2023:itransformer}, PatchTST~\cite{nie:2022patchtst}, FEDformer~\cite{zhou:2022fedformer}, and Autoformer~\cite{wu:2021autoformer}; and (iv) the MLP-based model, DLinear~\cite{zeng:2023}.

\renewcommand{\thefootnote}{\arabic{footnote}}
\paragraph{Implementation Details.}
To ensure the fairness of experiments, all baseline models are configured according to the settings in their official open-source repositories and evaluated using a unified evaluation framework, TSLib \footnote{\url{https://github.com/thuml/Time-Series-Library}}. All experiments are implemented based on PyTorch~\cite{paszke:2019pytorch} and executed on an NVIDIA A100-80G GPU. The default optimizer for the experiments is Adam~\cite{diederik2014:adam}, and each experiment is repeated three times, with the average results reported. 

\subsection{Performance Comparison}
Table \ref{table:2} presents the experimental results for long-term multivariate forecasting, highlighting the substantial advantages of our model. Specifically, compared to LLM-based methods, our model reduces MSE and MAE by 13.19\% and 5.35\%, respectively. When compared to Multi-scale models, the reductions are 24.99\% and 15.14\%, respectively. Relative to the DLinear model, MSE and MAE decrease by 22.73\% and 13.35\%. Furthermore, our model exhibits even more significant improvements over other transformer-based models. Notably, even when compared with SOTA models such as TIME-LLM~\cite{jin:2023time-llm} and PatchTST~\cite{nie:2022patchtst}, our model consistently delivers superior performance. These results strongly demonstrate the robustness and superiority of our approach across various datasets.

\begin{table*}[t]
\renewcommand\arraystretch{1.1}
\centering
\resizebox{0.99\textwidth}{!}{%
\begin{tabular}{cc|cc|cc|cc|cc|cc|cc|cc|cc|cc|cc|cc}
\toprule
\multicolumn{2}{c|}{
  \begin{tabular}{@{}c@{}}
    Method
  \end{tabular}
}
& \multicolumn{2}{c|}{
  \begin{tabular}{@{}c@{}}
    AdaMixT\\(\textbf{Ours})
  \end{tabular}
}
& \multicolumn{2}{c|}{
  \begin{tabular}{@{}c@{}}
    TIME-LLM\\(2023)
  \end{tabular}
}
& \multicolumn{2}{c|}{
  \begin{tabular}{@{}c@{}}
    GPT4TS\\(2023)
  \end{tabular}
}
& \multicolumn{2}{c|}{
  \begin{tabular}{@{}c@{}}
    TimeMixer\\(2024)
  \end{tabular}
}
& \multicolumn{2}{c|}{
  \begin{tabular}{@{}c@{}}
    MICN\\(2023)
  \end{tabular}
}
& \multicolumn{2}{c|}{
  \begin{tabular}{@{}c@{}}
    TimesNet\\(2022)
  \end{tabular}
}
& \multicolumn{2}{c|}{
  \begin{tabular}{@{}c@{}}
    iTransformer\\(2023)
  \end{tabular}
}
& \multicolumn{2}{c|}{
  \begin{tabular}{@{}c@{}}
    PatchTST\\(2022)
  \end{tabular}
}
& \multicolumn{2}{c|}{
  \begin{tabular}{@{}c@{}}
    FEDformer\\(2022)
  \end{tabular}
}
& \multicolumn{2}{c|}{
  \begin{tabular}{@{}c@{}}
    Autoformer\\(2021)
  \end{tabular}
}
& \multicolumn{2}{c}{
  \begin{tabular}{@{}c@{}}
    DLinear\\(2023)
  \end{tabular}
} \\
\midrule
\multicolumn{2}{c|}{Metric} 
& MSE & MAE 
& MSE & MAE 
& MSE & MAE 
& MSE & MAE
& MSE & MAE
& MSE & MAE
& MSE & MAE
& MSE & MAE
& MSE & MAE
& MSE & MAE
& MSE & MAE \\ 
\midrule

\multicolumn{1}{c|}{\multirow{5}{*}{\rotatebox{90}{Weather}}}
& 96 
& \textbf{0.145} & \textbf{0.196}
& \underline{0.149} & 0.200
& 0.162 & 0.212
& 0.161 & 0.209
& 0.161 & 0.229
& 0.172 & 0.220
& 0.175 & 0.216
& \underline{0.149} & \underline{0.198}
& 0.238 & 0.314
& 0.249 & 0.329
& 0.176 & 0.238 \\

\multicolumn{1}{c|}{} 
& 192 
& \textbf{0.190} & \textbf{0.238}
& \underline{0.193} & 0.243
& 0.204 & 0.248
& 0.207 & 0.250
& 0.220 & 0.281
& 0.219 & 0.261
& 0.225 & 0.258
& 0.194 & \underline{0.241}
& 0.275 & 0.329
& 0.325 & 0.370
& 0.218 & 0.277 \\

\multicolumn{1}{c|}{} 
& 336 
& \textbf{0.243} & \textbf{0.279}
& \textbf{0.243} & 0.284
& 0.254 & 0.286
& 0.264 & 0.292
& 0.278 & 0.331
& 0.280 & 0.306
& 0.280 & 0.298
& \underline{0.245} & \underline{0.282}
& 0.339 & 0.377
& 0.351 & 0.391
& 0.262 & 0.313 \\

\multicolumn{1}{c|}{} 
& 720 
& \textbf{0.310} & \textbf{0.332}
& 0.315 & 0.336
& 0.326 & 0.337
& 0.344 & 0.343
& \underline{0.311} & 0.356
& 0.365 & 0.359
& 0.361 & 0.351
& 0.314 & \underline{0.334}
& 0.389 & 0.409
& 0.415 & 0.426
& 0.327 & 0.367 \\

\multicolumn{1}{c|}{} 
& Avg 
& \textbf{0.222} & \textbf{0.261}
& \underline{0.225} & 0.266
& 0.237 & 0.271
& 0.244 & 0.274
& 0.243 & 0.299
& 0.259 & 0.287
& 0.260 & 0.281
& 0.226 & \underline{0.264}
& 0.310 & 0.357
& 0.335 & 0.379
& 0.246 & 0.299 \\ 
\midrule

\multicolumn{1}{c|}{\multirow{5}{*}{\rotatebox{90}{Traffic}}} 
& 96 
& \textbf{0.358} & \textbf{0.248}
& 0.376 & 0.280
& 0.388 & 0.282
& 0.466 & 0.293
& 0.519 & 0.309
& 0.593 & 0.321
& 0.394 & 0.269
& \underline{0.360} & \underline{0.249}
& 0.576 & 0.359
& 0.597 & 0.371
& 0.413 & 0.288 \\

\multicolumn{1}{c|}{} 
& 192 
& \textbf{0.378} & \textbf{0.254}
& 0.397 & 0.294
& 0.407 & 0.290
& 0.507 & 0.301
& 0.537 & 0.315
& 0.617 & 0.336
& 0.412 & 0.277
& \underline{0.379} & \underline{0.256}
& 0.610 & 0.380
& 0.607 & 0.382
& 0.423 & 0.287 \\

\multicolumn{1}{c|}{} 
& 336 
& \textbf{0.390} & \textbf{0.263}
& 0.420 & 0.311
& 0.412 & 0.294
& 0.525 & 0.309
& 0.534 & 0.313
& 0.629 & 0.336
& 0.425 & 0.283
& \underline{0.392} & \underline{0.264}
& 0.608 & 0.375
& 0.623 & 0.387
& 0.438 & 0.300 \\

\multicolumn{1}{c|}{} 
& 720 
& \textbf{0.428} & \underline{0.287}
& 0.448 & 0.326
& 0.450 & 0.312
& 0.552 & 0.325
& 0.577 & 0.325
& 0.640 & 0.350
& 0.460 & 0.301
& \underline{0.432} & \textbf{0.286}
& 0.621 & 0.375
& 0.639 & 0.395
& 0.466 & 0.315 \\

\multicolumn{1}{c|}{} 
& Avg 
& \textbf{0.389} & \textbf{0.263}
& 0.410 & 0.303
& 0.414 & 0.295
& 0.513 & 0.307
& 0.542 & 0.316
& 0.620 & 0.336
& 0.423 & 0.283
& \underline{0.391} & \underline{0.264}
& 0.604 & 0.372
& 0.617 & 0.384
& 0.435 & 0.298 \\ 
\midrule

\multicolumn{1}{c|}{\multirow{5}{*}{\rotatebox{90}{Electricity}}}
& 96 
& \textbf{0.118} & \textbf{0.214}
& 0.137 & 0.244
& 0.139 & 0.238
& \underline{0.120} & \underline{0.215}
& 0.164 & 0.269
& 0.168 & 0.272
& 0.148 & 0.240
& 0.129 & 0.222
& 0.186 & 0.302
& 0.196 & 0.313
& 0.141 & 0.240 \\

\multicolumn{1}{c|}{} 
& 192 
& \textbf{0.146} & \textbf{0.237}
& 0.158 & 0.266
& 0.153 & 0.251
& 0.170 & 0.261
& 0.177 & 0.285
& 0.184 & 0.289
& 0.164 & 0.256
& \underline{0.147} & \underline{0.240}
& 0.197 & 0.311
& 0.211 & 0.324
& 0.158 & 0.260 \\

\multicolumn{1}{c|}{} 
& 336 
& \textbf{0.160} & \textbf{0.258}
& 0.183 & 0.292
& 0.169 & 0.266
& 0.187 & 0.278
& 0.193 & 0.304
& 0.198 & 0.300
& 0.178 & 0.271
& \underline{0.163} & \underline{0.259}
& 0.213 & 0.328
& 0.214 & 0.327
& 0.171 & 0.271 \\

\multicolumn{1}{c|}{} 
& 720 
& \textbf{0.194} & \textbf{0.289}
& 0.247 & 0.348
& 0.206 & 0.297
& 0.228 & 0.313
& 0.212 & 0.321
& 0.220 & 0.320
& 0.211 & 0.300
& \underline{0.197} & \underline{0.290}
& 0.233 & 0.344
& 0.236 & 0.342
& 0.206 & 0.304 \\

\multicolumn{1}{c|}{} 
& Avg 
& \textbf{0.155} & \textbf{0.250}
& 0.181 & 0.288
& 0.167 & 0.263
& 0.176 & 0.267
& 0.187 & 0.295
& 0.193 & 0.295
& 0.175 & 0.267
& \underline{0.159} & \underline{0.253}
& 0.207 & 0.321
& 0.214 & 0.327
& 0.169 & 0.269 \\ 
\midrule

\multicolumn{1}{c|}{\multirow{5}{*}{\rotatebox{90}{ILI}}} 
& 24 
& 1.384 & \underline{0.757}
& 1.708 & 0.765
& 2.063 & 0.881
& \underline{1.358} & 0.763
& 2.684 & 1.112
& 2.317 & 0.934
& 1.638 & 0.831
& \textbf{1.319} & \textbf{0.754}
& 2.624 & 1.095
& 2.906 & 1.182
& 1.964 & 0.975 \\

\multicolumn{1}{c|}{} 
& 36 
& \textbf{1.300} & \textbf{0.755}
& 1.634 & \underline{0.781}
& 1.868 & 0.892
& 1.432 & 0.826
& 2.667 & 1.068
& 1.972 & 0.920
& 1.742 & 0.879
& \underline{1.430} & 0.834
& 2.516 & 1.021
& 2.585 & 1.038
& 2.080 & 0.998 \\

\multicolumn{1}{c|}{} 
& 48 
& \textbf{1.475} & \underline{0.793}
& 1.597 & \textbf{0.769}
& 1.790 & 0.884
& \underline{1.551} & 0.814
& 2.558 & 1.052
& 2.238 & 0.940
& 1.826 & 0.932
& 1.553 & 0.815
& 2.505 & 1.041
& 3.024 & 1.145
& 2.163 & 1.043 \\

\multicolumn{1}{c|}{} 
& 60 
& \textbf{1.460} & 0.821
& 1.565 & \textbf{0.754}
& 1.979 & 0.957
& 1.614 & 1.827
& 2.747 & 1.110
& 2.027 & 0.928
& 1.954 & 0.973
& \underline{1.470} & \underline{0.788}
& 2.742 & 1.122
& 2.761 & 1.114
& 2.396 & 1.112 \\

\multicolumn{1}{c|}{} 
& Avg 
& \textbf{1.405} & \underline{0.782}
& 1.626 & \textbf{0.767}
& 1.925 & 0.904
& 1.489 & 1.058
& 2.664 & 1.086
& 2.139 & 0.931
& 1.790 & 0.904
& \underline{1.443} & 0.798
& 2.597 & 1.070
& 2.819 & 1.120
& 2.151 & 1.032 \\ 
\midrule

\multicolumn{1}{c|}{\multirow{5}{*}{\rotatebox{90}{ETTh1}}} 
& 96 
& \textbf{0.360} & \textbf{0.393}
& 0.398 & 0.414
& 0.376 & \underline{0.397}
& 0.381 & 0.398
& 0.421 & 0.431
& 0.384 & 0.402
& 0.386 & 0.405
& \underline{0.370} & 0.399
& 0.376 & 0.415
& 0.435 & 0.446
& 0.422 & 0.448 \\

\multicolumn{1}{c|}{} 
& 192 
& \textbf{0.398} & \textbf{0.418}
& 0.442 & 0.440
& 0.416 & \textbf{0.418}
& 0.442 & 0.430
& 0.474 & 0.487
& 0.436 & 0.429
& 0.441 & 0.436
& \underline{0.413} & \underline{0.421}
& 0.423 & 0.446
& 0.456 & 0.457
& 0.419 & 0.430 \\

\multicolumn{1}{c|}{} 
& 336 
& \textbf{0.398} & \textbf{0.427}
& 0.456 & 0.450
& 0.442 & \underline{0.433}
& 0.501 & 0.460
& 0.770 & 0.672
& 0.521 & 0.500
& 0.491 & 0.461
& \underline{0.422} & 0.436
& 0.444 & 0.462
& 0.486 & 0.487
& 0.460 & 0.462 \\

\multicolumn{1}{c|}{} 
& 720 
& \underline{0.453} & \underline{0.465}
& 0.602 & 0.545
& 0.477 & \textbf{0.456}
& 0.544 & 0.505
& 0.770 & 0.672
& 0.493 & 0.505
& 0.509 & 0.494
& \textbf{0.447} & 0.466
& 0.469 & 0.492
& 0.515 & 0.517
& 0.521 & 0.500 \\

\multicolumn{1}{c|}{} 
& Avg 
& \textbf{0.402} & \textbf{0.426}
& 0.475 & 0.462
& 0.428 & \textbf{0.426}
& 0.467 & 0.448
& 0.559 & 0.535
& 0.458 & 0.450
& 0.457 & 0.449
& \underline{0.413} & \underline{0.431}
& 0.428 & 0.454
& 0.473 & 0.477
& 0.449 & 0.461 \\ 
\midrule

\multicolumn{1}{c|}{\multirow{5}{*}{\rotatebox{90}{ETTh2}}} 
& 96 
& \textbf{0.260} & \textbf{0.328}
& 0.309 & 0.362
& 0.285 & 0.342
& 0.288 & 0.340
& 0.299 & 0.364
& 0.340 & 0.374
& 0.300 & 0.350
& \underline{0.274} & \underline{0.336}
& 0.332 & 0.374
& 0.332 & 0.368
& 0.279 & 0.344 \\

\multicolumn{1}{c|}{} 
& 192 
& \textbf{0.306} & \textbf{0.370}
& 0.362 & 0.395
& 0.354 & 0.389
& 0.391 & 0.403
& 0.441 & 0.454
& 0.402 & 0.414
& 0.382 & 0.400
& \underline{0.339} & \underline{0.379}
& 0.407 & 0.446
& 0.426 & 0.434
& 0.361 & 0.401 \\

\multicolumn{1}{c|}{} 
& 336 
& \textbf{0.306} & \textbf{0.372}
& 0.376 & 0.409
& 0.373 & 0.407
& 0.422 & 0.427
& 0.654 & 0.567
& 0.452 & 0.452
& 0.424 & 0.432
& \underline{0.329} & \underline{0.380}
& 0.400 & 0.447
& 0.477 & 0.479
& 0.466 & 0.473 \\

\multicolumn{1}{c|}{} 
& 720 
& \textbf{0.376} & \textbf{0.418}
& 0.405 & 0.436
& 0.406 & 0.441
& 0.442 & 0.451
& 0.956 & 0.716
& 0.462 & 0.468
& 0.426 & 0.445
& \underline{0.379} & \underline{0.422}
& 0.412 & 0.469
& 0.453 & 0.490
& 0.398 & 0.417 \\

\multicolumn{1}{c|}{} 
& Avg 
& \textbf{0.312} & \textbf{0.372}
& 0.363 & 0.401
& 0.355 & 0.395
& 0.386 & 0.405
& 0.588 & 0.525
& 0.414 & 0.427
& 0.383 & 0.407
& \underline{0.330} & \underline{0.379}
& 0.388 & 0.434
& 0.422 & 0.443
& 0.351 & 0.409 \\ 
\midrule

\multicolumn{1}{c|}{\multirow{5}{*}{\rotatebox{90}{ETTm1}}} 
& 96 
& \textbf{0.288} & \textbf{0.341}
& \textbf{0.288} & 0.343
& 0.292 & 0.346
& 0.317 & 0.358
& 0.316 & 0.362
& 0.338 & 0.375
& 0.341 & 0.376
& \underline{0.290} & \underline{0.342}
& 0.326 & 0.390
& 0.510 & 0.492
& 0.303 & 0.346 \\

\multicolumn{1}{c|}{} 
& 192 
& \textbf{0.328} & \underline{0.369}
& 0.347 & 0.378
& \underline{0.332} & 0.372
& 0.367 & 0.387
& 0.363 & 0.390
& 0.374 & 0.387
& 0.381 & 0.395
& \underline{0.332} & \underline{0.369}
& 0.365 & 0.415
& 0.514 & 0.495
& 0.338 & \textbf{0.367} \\

\multicolumn{1}{c|}{} 
& 336 
& \textbf{0.359} & \textbf{0.388}
& 0.368 & 0.394
& \underline{0.366} & 0.394
& 0.388 & 0.402
& 0.408 & 0.426
& 0.410 & 0.411
& 0.419 & 0.419
& \underline{0.366} & \underline{0.392}
& 0.392 & 0.425
& 0.510 & 0.492
& 0.375 & 0.393 \\

\multicolumn{1}{c|}{} 
& 720 
& \textbf{0.415} & \textbf{0.418}
& 0.421 & 0.423
& 0.417 & 0.421
& 0.454 & 0.443
& 0.481 & 0.476
& 0.478 & 0.450
& 0.486 & 0.456
& \underline{0.416} & \underline{0.420}
& 0.446 & 0.458
& 0.527 & 0.493
& 0.427 & 0.422 \\

\multicolumn{1}{c|}{} 
& Avg 
& \textbf{0.348} & \textbf{0.379}
& 0.356 & 0.385
& 0.352 & 0.383
& 0.382 & 0.398
& 0.392 & 0.414
& 0.400 & 0.406
& 0.407 & 0.412
& \underline{0.351} & \underline{0.381}
& 0.382 & 0.422
& 0.515 & 0.493
& 0.361 & 0.382 \\ 
\midrule

\multicolumn{1}{c|}{\multirow{5}{*}{\rotatebox{90}{ETTm2}}} 
& 96 
& \textbf{0.163} & \textbf{0.252}
& 0.168 & 0.257
& 0.173 & 0.262
& 0.175 & 0.259
& 0.179 & 0.275
& 0.187 & 0.267
& 0.184 & 0.267
& \underline{0.165} & \underline{0.255}
& 0.180 & 0.271
& 0.205 & 0.293
& \underline{0.165} & 0.257 \\

\multicolumn{1}{c|}{} 
& 192 
& \textbf{0.217} & \textbf{0.290}
& \underline{0.219} & 0.293
& 0.229 & 0.301
& 0.237 & 0.299
& 0.307 & 0.376
& 0.249 & 0.309
& 0.253 & 0.312
& 0.220 & \underline{0.292}
& 0.252 & 0.318
& 0.278 & 0.336
& 0.227 & 0.307 \\

\multicolumn{1}{c|}{} 
& 336 
& \textbf{0.272} & \underline{0.330}
& 0.275 & 0.332
& 0.286 & 0.341
& 0.296 & 0.338
& 0.325 & 0.388
& 0.321 & 0.351
& 0.315 & 0.352
& \underline{0.274} & \textbf{0.329}
& 0.324 & 0.364
& 0.343 & 0.379
& 0.285 & 0.342 \\

\multicolumn{1}{c|}{} 
& 720 
& \textbf{0.361} & \underline{0.383}
& 0.367 & \textbf{0.335}
& 0.378 & 0.401
& 0.393 & 0.395
& 0.502 & 0.490
& 0.408 & 0.403
& 0.412 & 0.406
& \underline{0.362} & 0.385
& 0.410 & 0.420
& 0.414 & 0.419
& 0.398 & 0.417 \\

\multicolumn{1}{c|}{} 
& Avg 
& \textbf{0.253} & \underline{0.314}
& 0.257 & \textbf{0.304}
& 0.267 & 0.326
& 0.275 & 0.323
& 0.328 & 0.382
& 0.291 & 0.333
& 0.291 & 0.334
& \underline{0.255} & 0.315
& 0.292 & 0.343
& 0.310 & 0.357
& 0.269 & 0.331 \\ 
\midrule

\multicolumn{2}{c|}{$1^{st}$ Count} 
& \textbf{38} & \textbf{30} 
& \underline{2} & \underline{5} 
& 0 & 3 
& 0 & 0 
& 0 & 0 
& 0 & 0
& 0 & 0
& \underline{2} & 3
& 0 & 0
& 0 & 0
& 0 & 1 \\ 
\bottomrule
\end{tabular}
}
\caption{Long-term forecasting results on different datasets. ``Avg'' is the average of all considered prediction lengths. Lower MSE/MAE indicates better performance. We use prediction lengths $K \in \{24, 36, 48, 60\}$ for ILI and $K \in \{96, 192, 336, 720\}$ for the others. The best and second-best results are marked in \textbf{bold} and \underline{underlined}, respectively.
}
\label{table:2}
\end{table*}

\subsection{Model Analysis}

\paragraph{Ablation Study.} To comprehensively evaluate the contributions of the key components—GPM, DSM, and AWGN—to the overall performance of the AdaMixT, we conducted a series of ablation experiments. These experiments systematically removed each component, allowing us to compare the performance of the resulting model variants and quantify the impact of each element. 

\begin{figure}[t]
\centering
\subfigure
{\includegraphics[width=0.9\hsize]
{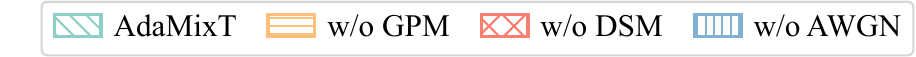}} \\
\setcounter{subfigure}{0}
\vspace{-0.2em}
\subfigure{
\begin{minipage}[t]{0.48\linewidth}
{\includegraphics[width=1\hsize]
{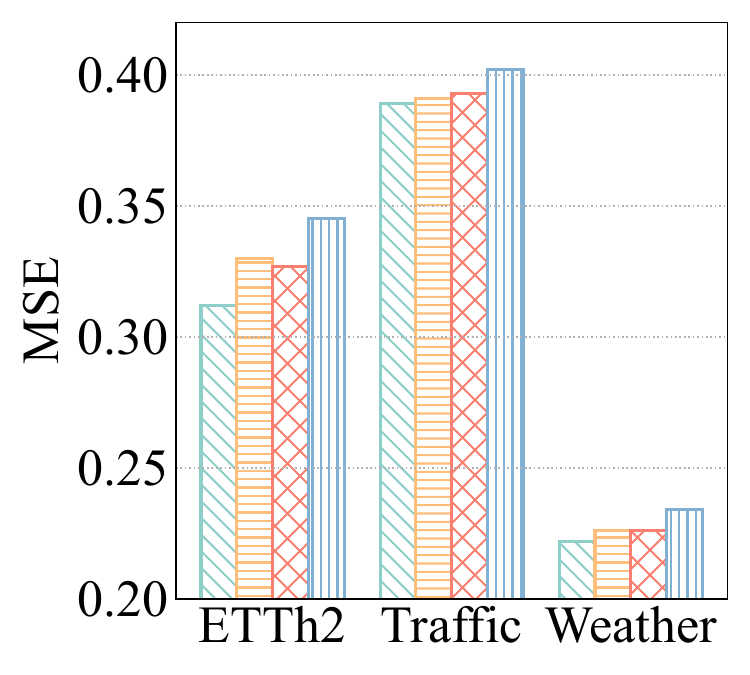}}
\end{minipage}}
\hspace{-1em}
\subfigure{
\begin{minipage}[t]{0.48\linewidth}
{\includegraphics[width=1\hsize]
{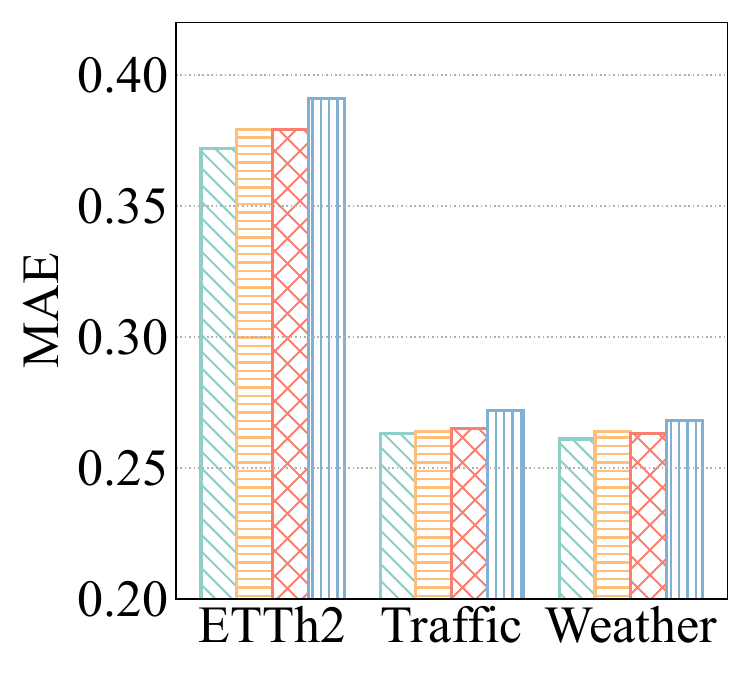}} 
\end{minipage}}
\vspace{-1em}
\caption{Average MSE and MAE comparison with different model variants on ETTh2, Traffic, and Weather datasets.}
\label{fig:ablation}
\end{figure}

\begin{figure}[t]
\centering
\subfigure
{\includegraphics[width=0.65\hsize]
{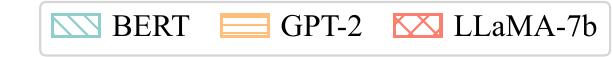}} \\
\setcounter{subfigure}{0}
\vspace{-0.2em}
\subfigure{
\begin{minipage}[t]{0.485\linewidth}
{\includegraphics[width=1\hsize]
{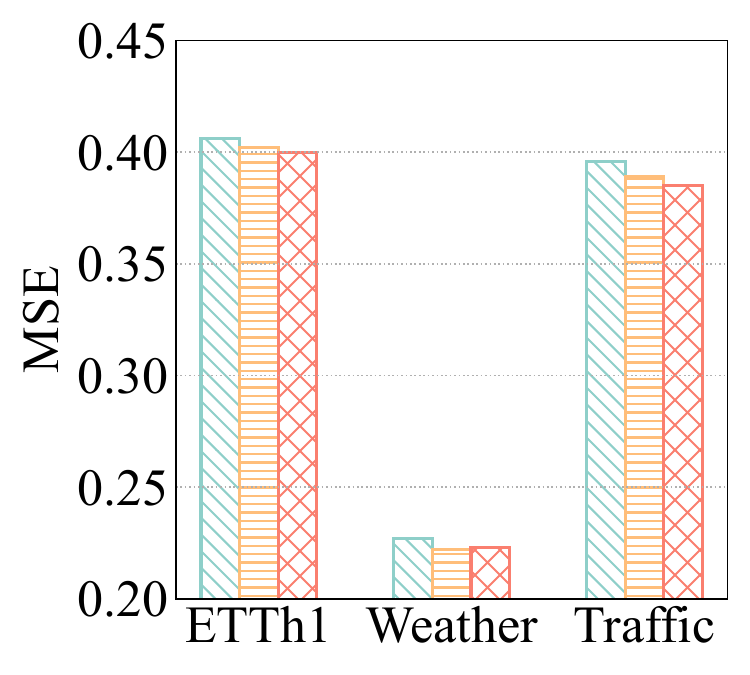}}
\end{minipage}}
\hspace{-1em}
\subfigure{
\begin{minipage}[t]{0.485\linewidth}
{\includegraphics[width=1\hsize]
{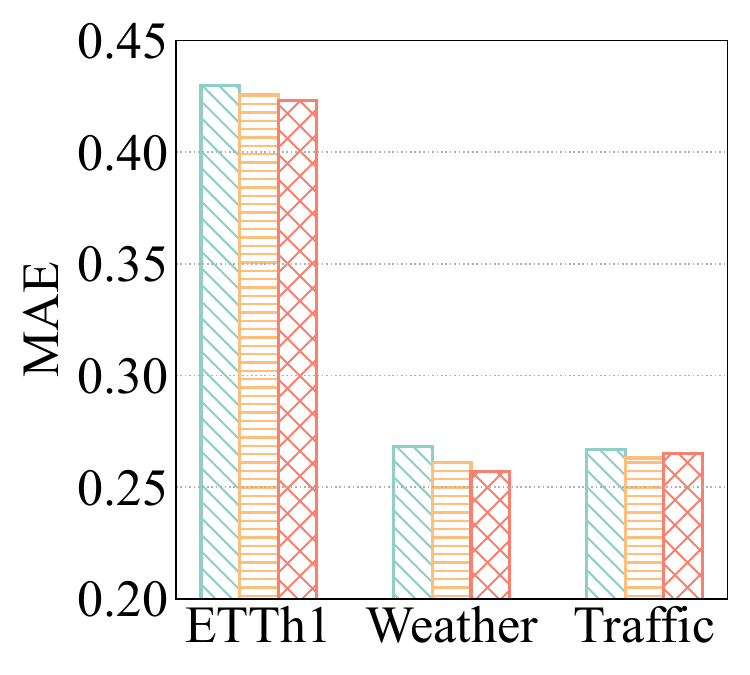}} 
\end{minipage}}
\vspace{-1em}
\caption{Multivariate long-term forecasting results with different pretrained models in AdaMixT.}
\vspace{-0.5em}
\label{fig:models}
\end{figure}

As shown in Figure~\ref{fig:ablation}, the ablation study results demonstrate that the complete AdaMixT model achieves the best performance on the ETTh2, Weather, and Traffic datasets, while removing any single module leads to a decrease in prediction accuracy. For example, on the ETTh2 dataset, removing GPM and DSM results in performance degradation, highlighting the critical role of GPM in extracting general feature representations and the significant value of DSM in modeling temporal features. Notably, the most substantial performance drop is observed when AWGN is removed, with both MSE and MAE significantly higher than other variants. This finding underscores the pivotal role of AWGN in multi-scale feature fusion and its importance in improving prediction accuracy. Similar trends are observed on the Weather and Traffic datasets, further confirming the importance of each module in enhancing model performance.

\paragraph{Results with Different Models.}
In the main experiments, AdaMixT utilizes GPT-2~\cite{radford:2019gpt2} as GPM. To further validate the effectiveness of the model, we also perform comparative experiments using BERT and LLaMA-7B. The experimental results in Figure~\ref{fig:models} indicate that the performance differences among BERT, GPT-2, and LLaMA-7B are minimal on the ETTh1, Weather, and Traffic datasets. This finding demonstrates the high robustness of AdaMixT in selecting GPMs, as it consistently delivers stable performance across various pretrained models.

\begin{figure*}[t]
\centering
\subfigure{
\begin{minipage}[t]{0.3\linewidth}
{\includegraphics[width=1\hsize]
{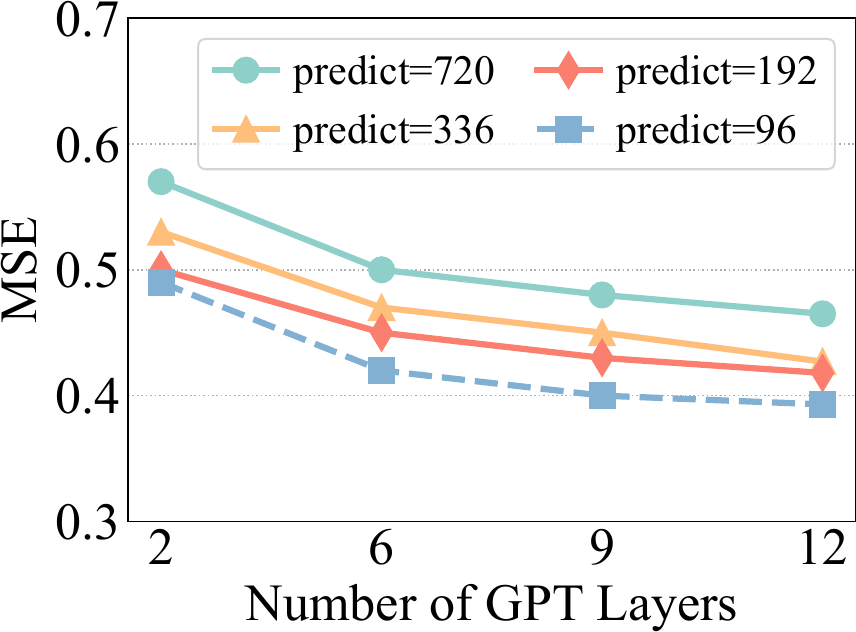}}
\end{minipage}}
\subfigure{
\begin{minipage}[t]{0.3\linewidth}
{\includegraphics[width=1\hsize]
{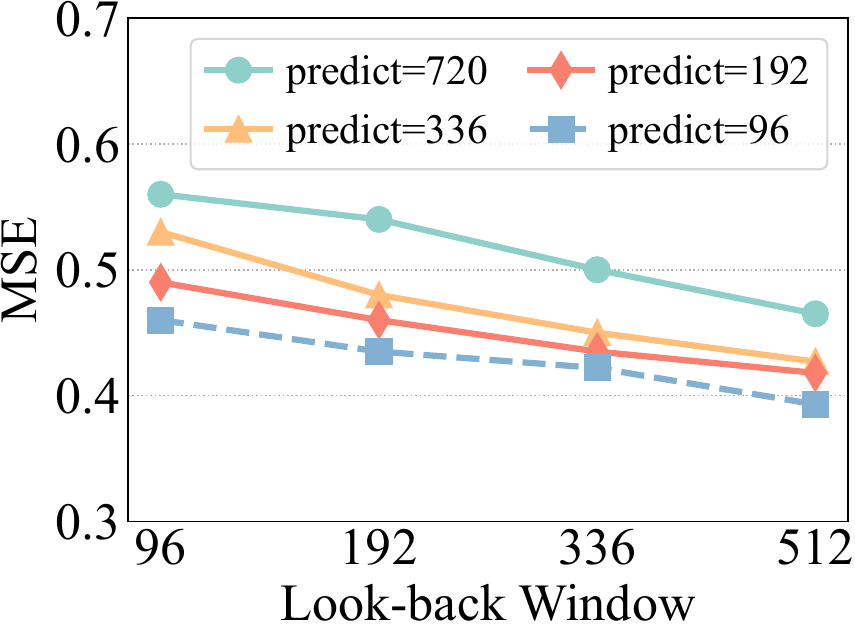}} 
\end{minipage}}
\subfigure{
\begin{minipage}[t]{0.3\linewidth}
{\includegraphics[width=1\hsize]
{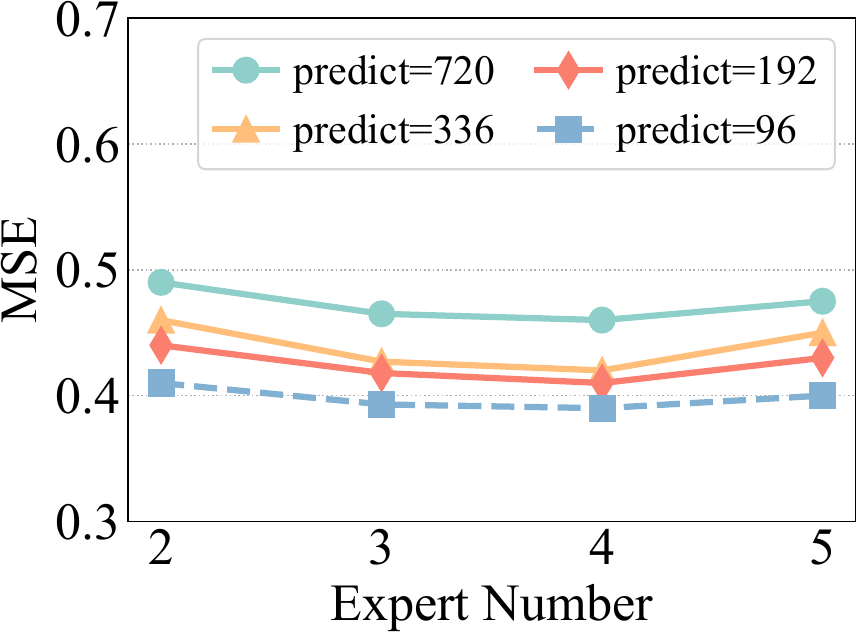}} 
\end{minipage}}
\vspace{-1em}
\caption{Sensitivity analysis of three key hyperparameters on the ETTh1 dataset.}
\label{fig:Sensitivity1}
\end{figure*}

\paragraph{Hyperparameter Sensitivity.}
We conducte a sensitivity analysis on 3 key hyperparameters, including the number of layers in the backbone model, the look-back window, and the number of experts. Results are shown in Figure~\ref{fig:Sensitivity1}. Based on the results, we summarize the following conclusions:
\begin{itemize}[leftmargin=0.5cm]
\item \textbf{Backbone Layers}: The number of layers in the backbone is positively correlated with the predictive performance. This indicates that, even after fusion, GPM retain favorable scaling laws with respect to layer depth, which positively influences model performance.
\item \textbf{Look-back Window}: The Look-back window directly affects prediction accuracy, especially at extended forecasting horizons. This observation aligns with the patterns found in traditional models, demonstrating that longer historical inputs can effectively enhance performance.
\item \textbf{Number of Experts}: The experimental results indicate that increasing the number of expert models can effectively capture features at different scales, thereby significantly improving the accuracy of time series forecasting. This result validates the effectiveness of the feature fusion mechanism in the AdaMixT. However, caution should be taken as the model may be prone to overfitting.
\end{itemize}

\begin{table}[!t]
\renewcommand\arraystretch{1.0}
\centering
\resizebox{0.45\textwidth}{!}{%
\begin{tabular}{c|c|cc|cc|cc}
\toprule
\multicolumn{2}{c|}{Method} & \multicolumn{2}{c|}{AdaMixT(1, 1/2)} & \multicolumn{2}{c|}{AdaMixT(1, 1)} & \multicolumn{2}{c}{AdaMixT(1, 2)}\\ 
\midrule
\multicolumn{2}{c|}{Metric} & MSE & MAE & MSE & MAE & MSE & MAE\\ 
\midrule
\multirow{5}{*}{\rotatebox{90}{ILI}} 
& 24 & 1.702 & 0.854 & 1.630 & 0.824 & \textbf{1.384} & \textbf{0.757}\\
& 36 & 1.753 & 0.862 & 1.509 & 0.816 & \textbf{1.300} & \textbf{0.755}\\
& 48 & 1.737 & 0.860 & 2.023 & 0.941 & \textbf{1.475} & \textbf{0.793}\\
& 60 & 1.560 & 0.839 & 1.630 & 0.891 & \textbf{1.460} & \textbf{0.821} \\
\midrule
\multirow{5}{*}{\rotatebox{90}{ETTh1}} 
& 96 & \textbf{0.360} & \textbf{0.393} & 0.366 & 0.399 & 0.381 & 0.408\\
& 192 & \textbf{0.398} & \textbf{0.418} & 0.400 & 0.415 & 0.412 & 0.423\\
& 336 & \textbf{0.398} & \textbf{0.427} & 0.411 & 0.431 & 0.407 & 0.428\\
& 720 & \textbf{0.453} & \textbf{0.465} & 0.459 & 0.470 & 0.571 & 0.547\\
\bottomrule
\end{tabular}}%
\caption{Impact of different feature scales on prediction accuracy for ILI and ETTh1 datasets. The best results are marked in \textbf{bold}.}
\label{table:4}
\end{table}

\paragraph{Scale Factors Study.}
As shown in Algorithm~\ref{alg:AdaMixT}, the setting of scale factors determines the granularity of feature extraction. The selection of the appropriate parameters is critical to the accuracy of the prediction. To further investigate this, we conduct an analysis of the impact of different feature scale settings on prediction performance, using the ILI and ETTh1 datasets as examples. 

Through a cyclic study of these two datasets, we find that the cyclicity of ILI is significantly longer than that of ETTh1. Based on this observation, we employ two expert models in the experiment, with the scale factor for GPM set to 1, and the scale factor range for the DSM set to \{1/2, 1, 2\}. The results, as shown in Table \ref{table:4}, indicate that for datasets with longer periodicities, using larger scale factors improves prediction performance (and vice versa). This finding suggests that when selecting scale factors, the intrinsic characteristics of the time series should be considered to achieve optimal prediction results.

\begin{figure}[!t]
\centering
\includegraphics[width=1\linewidth]{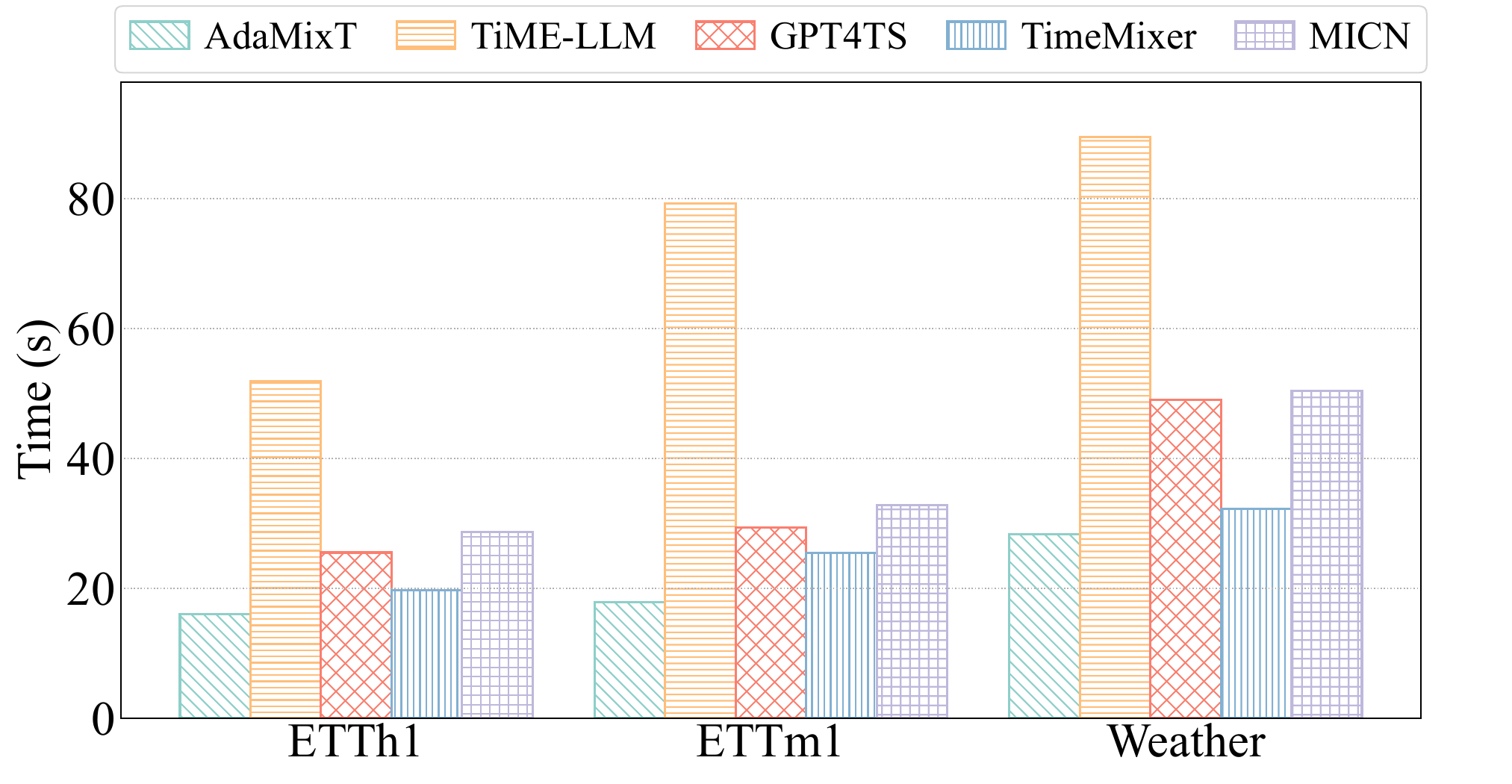}
\vspace{-1em}
\caption{Comparison of inference time for different models across ETTh1, ETTm1, and Weather datasets.}
\label{fig:infertime}
\end{figure}

\paragraph{Inference Time Study.} In order to evaluate the practical applicability, we compare the inference time of AdaMixT with current similar methods. As shown in Figure~\ref{fig:infertime}, AdaMixT demonstrates superior inference time compared with other complex multi-scale feature fusion models and LLM-based models. This advantage is primarily attributed to the fact that AdaMixT does not require complex prompt generation, post-processing, or frequency domain transformations. 

\section{Conclusion and Future Works}
In this paper, we present AdaMixT, which is designed to address the limitations of existing methods in terms of generalizability and multi-scale feature fusion. AdaMixT incorporates three key innovations: Multi-scale Feature Extraction, Expert Pool, and Adaptive Weighted Gating Network, significantly enhancing the model’s performance. To the best of our knowledge, this work is the first to combine the general feature representation capability of GPM with the fine-grained modeling ability of DSM to construct the expert pool. Additionally, we design AWGN that evaluates the weights of different models, enabling effective fusion of multi-scale features. Extensive experimental results on multiple benchmark datasets demonstrate that our approach outperforms existing forecasting approaches.

Our model offers new insights into multi-scale feature fusion in time series analysis. In the future, we aim to further explore multi-scale fusion methods and strengthen the complementary strengths of GPM and DSM, which will be a crucial research direction moving forward.

\section*{Acknowledgments}
This work is supported by the National Key Laboratory under grant 241-HF-Z01-01. Guoliang Li is supported by the National Key R\&D Program of China (2023YFB4503600), NSF of China (62232009),  Zhongguancun Lab, Huawei, and Beijing National Research Center for Information Science and Technology (BNRist).

\bibliographystyle{named}
\bibliography{ijcai25}

\end{document}